# Challenges and Opportunities for Machine Learning Classification of Behavior and Mental State from Images

Peter Washington, Cezmi Onur Mutlu, Aaron Kline, Kelley Paskov, Nate Tyler Stockham, Brianna Chrisman, Nick Deveau, Mourya Surhabi, Nick Haber, Dennis P. Wall

*Stanford University*

## Abstract

Computer Vision (CV) classifiers which distinguish and detect nonverbal social human behavior and mental state can aid digital diagnostics and therapeutics for psychiatry and the behavioral sciences. While CV classifiers for traditional and structured classification tasks can be developed with standard machine learning pipelines for supervised learning consisting of data labeling, preprocessing, and training a convolutional neural network, there are several pain points which arise when attempting this process for behavioral phenotyping. Here, we discuss the challenges and corresponding opportunities in this space, including handling heterogeneous data, avoiding biased models, labeling massive and repetitive data sets, working with ambiguous or compound class labels, managing privacy concerns, creating appropriate representations, and personalizing models. We discuss current state-of-the-art research endeavors in CV such as data curation, data augmentation, crowdsourced labeling, active learning, reinforcement learning, generative models, representation learning, federated learning, and meta-learning. We highlight at least some of the machine learning advancements needed for imaging classifiers to detect human social cues successfully and reliably.

## Background

Machine Learning (ML), and in particular Computer Vision (CV), is increasingly used across many if not all sectors and industries, including but not limited to healthcare [62, 71], robotics [108], security systems [167], industrial settings [189], autonomous vehicles [128], biology [Danuser], and astronomy [21]. While theoretical ML techniques have been validated using benchmark datasets, these applications have been limited to unchallenging domains. Complex vision tasks require nontrivial advancements in ML methodologies, and these approaches often must be tailored to the domain of interest. Here, we focus on CV for a particularly challenging yet widely relevant domain: nonverbal social human behaviors and underlying mental states such as emotion evocation, engagement, eye contact, paying attention to a conversation partner, hand gestures, body language, aggression, distraction, hyperactivity, and other nonverbal communicative behaviors. Understanding such nonverbal communication through vision would naturally need to be a prerequisite for passing the Turing test [193], and such study is therefore of general interest to the artificial intelligence (AI) research community. However, realizing sensitive and specific CV models for recognizing and classifying complex human behavior is a particularly enigmatic and demanding task, and there is a plethora of difficulties which can make the established process of simply labeling datasets and training a convolutional neural network (CNN) on the resulting labels yield low performance.

Classifying and detecting human mental states, social cues, and behavior with automated methods is largely beneficial for the field of psychiatry and behavior sciences [91, 180, 242].

"Digital phenotyping" has been coined by Insel et al. as the use of digital devices to passively monitor human subjects and find correlations between device usage patterns and behavioral patterns [91, 150]. In addition to passive measurement of behavior, recent digital phenotyping efforts have broadened to include structured digital interventions ("active" rather than "passive") which measure and adapt to user behavior. This paradigm has been researched for digital diagnostics and therapeutics for autism [35, 37, 57, 60, 161, 169, 226], Attention-deficit/hyperactivity disorder [9, 47, 105-106], major depressive disorder [10, 77], multiple sclerosis [22, 88], and learning disabilities [68]. Images and videos recorded on smartphones, augmented or virtual reality wearables, and webcams can be rich sources of data for CV classifiers which detect information about an individual's emotional state [82, 169, 195-196], social engagement [176], gait [84, 136, 140], psychical activity [15, 134, 170], sleep and rest [6, 171], violence [143], intoxication [129, 156], and several other relevant features [12, 149]. Digital phenotyping with CV has been previously attempted for video surveillance of sleep [130], emotion recognition [16, 63, 81, 93, 107, 183, 243-245], quantifying eating behavior [87, 164, 182], sign language recognition [14, 18], surveillance to prevent suicide attempts [25, 112], detecting distracted driving [4, 13, 147, 191], and several other behaviors, and the list continues to rapidly expand.

Current results are incredibly promising, but the underlying ML methods require development before they can fully support these ambitious yet highly impactful endeavors. Quantifying social human behavior and mental state, with all its nuances and complexities, is one of the most challenging CV tasks to date. We focus this review on the challenges and opportunities in core ML methods required to translate CV-powered systems into commercial, clinical, and government settings. CV is a vast field which spans tasks such as scene reconstruction, object tracking, and segmentation, including in 3D contexts. For the purposes of this review, we focus solely on *classification* of human behaviors and social cues from images and video streams. While classification is arguably one of the most straightforward CV tasks, the challenges we outline impede development of practically useful models.

Common issues often arise from the complexity and heterogeneity of the underlying data, including noisy data streams, biased datasets, massive and repetitive data, subjective class labels, a vast array of possibilities for feature representations, privacy issues, and the need for model personalization. While much work and success has been achieved in ML for social behavior, the challenges we identify here have usually resulted in poor model performance by commercial standards, limiting widespread productization of such models. We first outline and discuss these challenges which plague CV researchers and practitioners. We then describe corresponding opportunities for developing CV for detecting social human behavior (Table 1). In particular, we describe opportunities for innovation to enable such classifiers to be translated out of research settings, including strategies for automatic labeling, manual feature engineering, representation learning, active learning, generative data augmentation, crowd worker selection, federated learning, meta-learning, and self-supervised learning.

| Challenge | Opportunities | | | | | | | |
|---|---|---|---|---|---|---|---|---|
| | Automatic Labeling | Manual Feature Engineering | Self Supervised Learning | Active Learning | Generative Data Augmentation | Crowd Worker Selection | Federated Learning | Meta Learning |
| Handling data heterogeneity | | | | X | X | | | |
| Labeling large, high-dimensional, and repetitive data | X | | | X | | X | | |
| Avoiding biased models | | X | X | X | X | X | | |
| Reconciling subjective classes | X | | | | | X | | |
| Creating appropriate representations | | X | X | | | | | |
| Addressing privacy concerns | X | X | X | | | X | X | |
| Personalizing models | | X | X | | X | | | X |

**Table 1.** Summary of the discussed challenges and opportunities for developing improved CV models of social human behavior. We outline this review accordingly.

## Challenges

We begin this review with an overview of the methodological challenges which hinder the development of high-performing CV models for complex domains such as social behavioral classification. We discuss challenges with handling heterogeneous data, labeling massive and high-dimensional data streams, avoiding biased models, reconciling subjective classes, creating appropriate representations, addressing privacy concerns, and personalizing models.

*Handling Data Heterogeneity*

Video data collected from camera streams such as webcams, security footage, autonomous vehicles, robots, and smartphone or handheld cameras are inherently noisy, especially when the data are collected in naturalistic settings. Precise behavioral data acquisition efforts can and have been run in controlled environments as parts of in-lab studies (e.g., [121, 151, 220, 224]), but classifiers trained using laboratory data do not often generalize to naturalistic settings [31, 236], since animal and human behavior in lab settings is well-documented to differ from in-the-wild environments [158]. Slight deviations from lab conditions can lead to unintended and detrimental overfitting as a natural implication of dataset bias, where machine learning models are biased towards a dataset's non-essential attributes [187, 237]. Therefore, practically useful datasets for high-dimensional complex discriminative tasks require large scale data collection efforts across a variety of contexts per class, most often in unmediated settings. To account for variation which is irrelevant to the discriminative task across data points and to minimize the inevitable resulting overfitting, concerted effort is required to both minimize overall noise and to equalize noise across classes and demographics [187-188, 237]. We discuss a multitude of solutions, including thoughtful data collection pipelines, feature engineering approaches, and data augmentation strategies.

*Labeling Massive, High Dimensional, Complex, and Repetitive Data Sets*

Digital phenotyping data streams are often massive due to the continuous nature of the data collection. CV data collection for computational psychiatry have come from naturalistic home videos recorded using the front-facing camera of the phone [96-99] as well as wearable devices [44-45, 81-82, 102, 195-196, 210-211] during gamified therapy sessions. With repeated use, such longitudinal interventions generate massive data streams which can be used as training data for behavioral classifiers. Given the infeasibility, in terms of both time and cost, of labeling all data with ground truth human annotations, researchers have confronted the issue from two angles: (1) using model uncertainty or other metrics to intelligently prioritize frames for labeling (a technique called "active learning"), and (2) automatically labeling frames using unsupervised and representational approaches. We discuss both classes of solutions.

*Avoiding Biased Models*

Biased ML plagues a plethora of human-centered applications of ML. Datasets of human behavior are particularly prone to biases from many aspects of the data acquisition pipeline, and the high dimensionality of imaging data tends to magnify these issues.

One of the worst situations that can occur is batch effects which differ across classes. For example, facial expressions are emoted differently across cultures [131, 166], and the interpretation of the expression label also differs across labelers due to cultural differences in emotion perception [61]. Any classifier should in principle perform equally for all groups it is developed to serve, and data should be collected in similar or identical settings for some groups. For example, let's say we are predicting behavioral symptoms of a particular psychiatric condition using CV. Because the United States has a larger White population than Black [72], it is possible that more datapoints will be collected for the White population than the Black population in a large-scale data collection effort. A first-pass solution is to simply balance the dataset by demographics such that each outcome variable has equal representation of all demographic groups. However, such a solution will not account for the reduced diversity of settings in the Black subgroup compared to the White. Even if balancing the number of data points, random selection of images will result in a greater variation in background setting and lighting, camera configurations, and subtleties of behavior in the White subgroup compared to the Black, thus resulting in a more robust classifier when evaluating on the White population. Such racial disparities in ML have been repeatedly documented [36, 73, 133]. Similar disparities have also been documented for gender-related attributes [29, 73, 159, 201].

While attempts to debias models can take place for attributes anticipated by engineers in advance, one of the largest difficulties with debiasing methods is that they often require understanding of the potential biases in advance. However, not all potential confounders can be explicitly enumerated, especially for images with complex nonlinear interactions between features and sensitive demographic attributes. Complex image datasets are especially prone to unanticipated confounders [187] and uneven attribute effects [118, 221] such as skin, hair, or eye color [113], gender-related phenotypes [29], clothing [38, 53], and other visual artefacts [186] which can be detected and visualized via pixel saliency maps [154-155]. The confounders can also be indescribable and result from an intricate combination of abstract features [49, 138].

There is much room for innovation in the development of bias detection and mitigation algorithms for images and videos which can work for sophisticated visual features.

Biases may also arise not due to the data itself but because of biases by the annotators of the data. If crowdsourcing is utilized, the biases of the crowd workers must be accounted for. Crowd workers are humans with their own preconceptions and predispositions, and these cognitive biases have been documented in crowdsourcing contexts [58]. Classifiers for human social behavior are again particularly susceptible to such biases due to the richness of information provided in the images coupled with the innate subjectivity of interpretating human social behavior.

Biases can even occur prior to the data collection process. The decision of which data to record and which classifiers to build can introduce some of the strongest sources of bias which magnify downstream. For example, data related to autism are collected based on diagnostic criteria, but these criteria are based on *assumed* deviations from normal social behavior, since data pertaining to the true norm were absent at the time of creation of the diagnostic criteria [123-124]. The increasing utility of big data analysis on large datasets can help mitigate qualitative selection biases by informing decisions about which classifiers should be built in the first place using data-driven decision making. Feature selection methods, for example, can help analyze large scale medical records to determine which classifiers should be built for detection of autism [1-3, 110, 114, 198, 208].

*Reconciling Subjective Classes*

A major challenge when building CV models, or ML models in general, for measuring human behavior is the subjectivity of the labels. Interpretation of human behavior is widely understood to be subjective, including language and communication [20], understanding of emotions [90], and qualification for a psychiatric diagnosis [28]. This inherent ambiguity creates a nuisance for ML model development, as a ground truth label is required for the training process. Related to the ambiguity of class membership is compound classes, where a particular label contains multiple correct and overlapping labels. In the case of affective computing, a face may be purely surprised, but it may be also "happily surprised", "fearfully surprised", or "angrily surprised" [55].

*Creating Appropriate Representations*

There are few standard feature representations for human social behaviors, and this is possibly the largest bottleneck in creating successful models. Advancements in manually engineered and learned feature representation methods will likely enable some of the most dramatic advancements in CV classifiers moving forward [19].

Representation learning, or feature learning, is the use of automated methods to learn feature extraction techniques which can generalize to a particular set of supervised tasks. A strong feature representation allows a model to "look for" the right things when making predictive decisions while ignoring inconsequential features which would lead to overfitting. The fundamental reason behind the success of CNNs in CV is their ability to learn such complex visual features automatically via data-driven kernel weights [5]. Representations learned during supervised learning in one domain can often transfer to other domains, with more similar source domains usually resulting in more effective target domain performance [233]. Strong representation learning approaches can minimize the number of human labels required for each image in the dataset [19].

*Addressing Privacy Concerns*

The training data for behavioral CV models are particularly sensitive to privacy concerns [212]. Data are often recorded in home settings and may include private content. For human subjects to feel comfortable sharing their images, videos, and audio streams with annotators and engineering teams, extra precautions must be put in place to garner maximum trust from the user. Images and audio are more sensitive than other data streams used in behavioral phenotyping such as sensor data, application usage, device usage, and even location data. The popularity of social media applications which are generally accepted to collect such data has bolstered public apprehension towards sharing their data in other contexts [92, 214].

One approach for trustworthy data sharing containing sensitive protected health information (PHI) is to modify the training data to obfuscate the identity of the individual. If a subject cannot possibly be identified when sharing their data, then they may be more likely to share with an external organization. Privacy modifications to training data have included local modifications like placing a box over the subject's face and pitch shifting the audio in the video [209] as well as global modifications such as applying blurring, Gaussian noise, or optical flow to the entirety of the video frame or region of interest [204]. The fundamental tradeoff is between the strength of privacy preservation and the degree of information loss. There is ample room for innovation in privacy preserving image alterations which clearly depict the behavior of interest while obscuring and PHI and ideally removing any irrelevant features as well. The best solutions will achieve both privacy preservation and minimization of information loss.

*Personalizing Models*

To increase model performance to commercial levels while also addressing privacy concerns, ML engineers may often consider personalizing models to each user. The key design goal of a personalized model is to overfit with respect to the subject's features without overfitting to the environment, camera quality, camera angle, or any other features not relevant to the classification task at hand. The feasibility of model personalization for digital phenotyping with CV is dependent on the desired production use case. If the training data used for personalization are acquired in the same environment as the production setting (e.g., the user's home) and at a consistent angle (e.g., from a security camera), then overfitting to the background is not detrimental and potentially even desired. If, on the other hand, the model needs to perform in a variety of settings, then data must be acquired for the subject in a vast variety of environments, which may be infeasible in certain contexts. The primary ML challenge when deploying personalized models is to enable the model to quickly optimize for the new target distribution with only a few examples.

**Opportunities**

Having covered the challenges for complex CV model development, we now review current methods which address these challenges. We then outline opportunities for future work and innovation. We examine the present state-of-the-art methodologies in automatic data labeling, manual feature engineering, feature representation learning including self-supervised learning, active learning, generative methods for data augmentation, label representation, crowd worker selection, federated learning, and meta-learning (Figure 1).

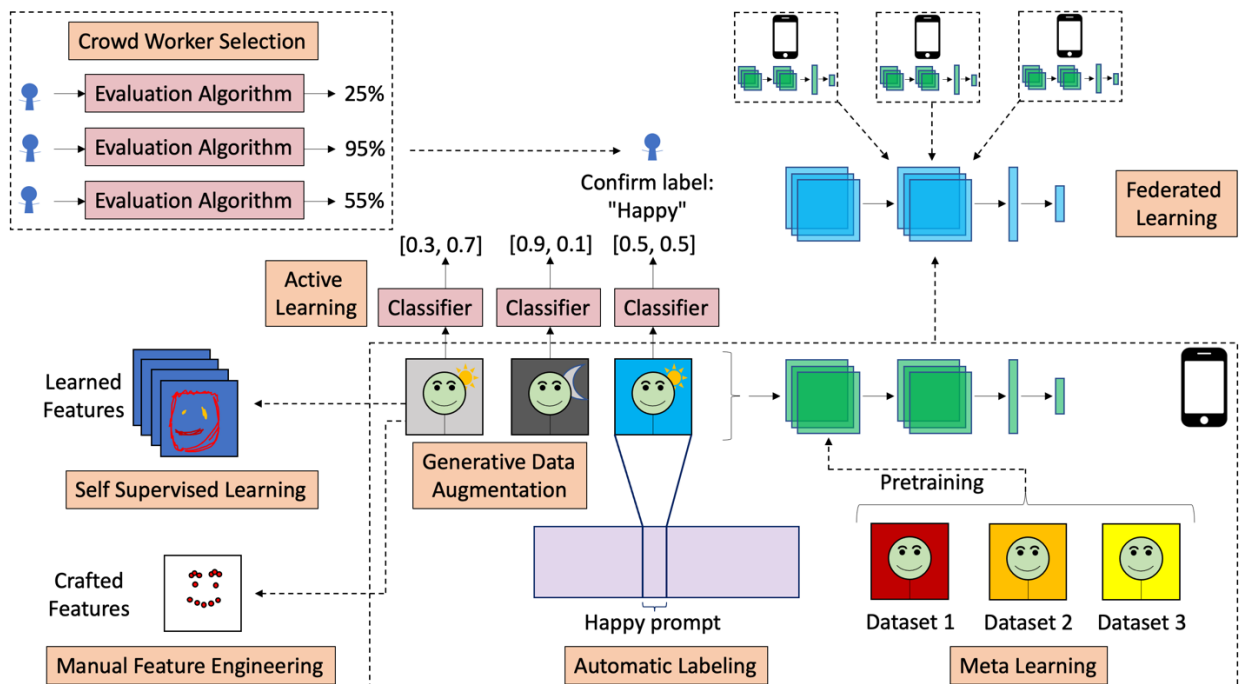

**Figure 1. Overview of opportunities in machine learning research to improve vision-based classification of social human behaviors.** The opportunities we discuss span the ML development lifecycle for CV models, and each challenge must be addressed before widespread production use of such models becomes feasible.

*Automatic Labeling*

Automatic data labeling schemes deployed in consumer devices can help alleviate some of the noise inherent in the data. For example, Kalantarian et al. developed a mobile therapeutic game for children with autism in which children interact with a parent or caregiver by acting out Charades prompts displayed on the phone while the parent holds the phone on their forehead [96-99]. Videos of the child are recorded through the front-facing camera of the device. Structure is imposed on the data through two mechanisms: the underlying game's inherent structure and metadata provided explicitly by the parent during the game. There is a one-to-one mapping between game prompts and data labels, and the time of a video during which a behavior prompt is displayed is likely to be enriched for the corresponding behavior. Parents validate these "automatic labels" by tilting the phone forward to indicate a correct prompt and backwards to indicate an incorrect prompt. Data collection and labeling mechanisms such as this, which impose structure in unstructured settings, can help to generate massively labeled imaging datasets by appreciably diminishing the volume of manual labels per unit of raw data.

Mining public data archives is another popular technique for curating large, labeled datasets [157, 177]. Benchmark datasets for some CV tasks have been generated at scale without requiring human labels through web scraping. Guo et al. acquired a dataset of celebrity faces by scraping web documents with multiple structured queries for each celebrity name [78]. Weakly labeled affective computing datasets have been constructed by leveraging tags on social media websites [229] and search engine queries [107]. Unlike faces and emotions, however, the majority of images and videos depicting social human behaviors do not contain metadata which can easily be scraped online. There is an opening for developers of interactive systems to embed automatic labeling mechanisms to enable the conception of usable data libraries.

*Manual Feature Engineering*

Oftentimes, datasets are generated in uncontrolled settings without classification in mind (for example, videos recorded from security cameras [39], in-the-wild datasets scraped from the Internet [135, 179], and streams collected from augmented reality experiences [7, 67]). With such heterogeneous raw data, the primary challenge for ML engineers is to computationally minimize the noise. Feature engineering is a family of techniques used for this purpose. The primary challenge in feature engineering is to extract features relevant to the classification task at hand without capturing irrelevant features which would lead to overfitting. These feature representations can either be hand-engineered or automatically learned. We first describe manual feature extraction techniques for human social behavior, focusing on facial emotion recognition as an archetypical social behavior with expansive exploration in prior literature.

Successful manual feature representations exist for several behaviors, including facial affect [51, 66, 122, 194] and physical activities [74, 145, 148]. Facial landmarks are a common feature representation for facial emotion recognition [80, 137, 142, 184]. The low-dimensional representation of facial landmarks allows for training with fewer images and reduces the risk for overfitting. The success of landmark-based classification, however, relies on the precision of the underlying landmark extraction algorithm. Several approaches exist for procurement of facial landmarks. Regression-based methods optimize for minimizing the difference between the predicted and manually annotated face shape [30, 34]. Regression models such as random forest regression voting [41, 225] and hierarchical regression with cascaded random ferns [238] have been used. Template fitting methods [40, 230] and deep learning approaches [181], for example using a cascade of CNNs [181], have also been evaluated. General-purpose feature representations are also used for facial emotion recognition, including Histograms of Oriented Gradients [42, 81] and Scale-Invariant Feature Transformations [174].

Manually engineered feature representations are optimized for existing publicly available datasets. For example, facial keypoint extractors find points around the subject's eyes, eyebrows, nose, mouth, and face outline [168, 173, 218, 232]. While such landmarks may be sufficient for recognition of basic universal emotions [59], which most public facial emotion datasets depict [103], this representation has not been evaluated on datasets containing nuanced facial expressions depicting "concerned", "skeptical", and "amused". Potential feature representations could include landmarks on the forehead, cheeks, and other underexplored areas of interest.

Common feature representations for generalized activity recognition which minimize noise not related to movement include dense optical flow [65], Lucas-Kanade optical flow [126], and pose estimation landmarks [43, 139, 190]. Popular CV libraries such as OpenCV [27] and OpenPose [32-33, 175, 213] ease the implementation barriers for utilizing these techniques, resulting in successes in simple activity recognition models which predict activities such as head banging [203], gait analysis [162] (including for dementia detection [235]), fall detection for seniors [89], yoga [223], cooking activities [148], gym activities [70], knee abduction in patients with osteoarthritis [24], and many others [79].

Specialized feature representations related to social behavior usually must be customized for a particular behavior or set of behaviors. For example, Wu et al. use extracted eye contours, the size of the iris, and pupil locations as features for an eye gaze direction model [219]. Facial muscle-related behaviors are frequently classified using facial action units as features. Facial action units are a taxonomy of facial movements described by the Facial Action Coding System [163]. Discriminative methods for extracting facial action units [17] have been used as input features for classification tasks such as facial pain displays [116], driver fatigue [197], confusion [216], and facial affect [185].

One approach for developing new behavioral classifiers is to manually construct domain-specific feature descriptors as we describe above. However, this paradigm has mostly become unfashionable in recent literature in favor of automatically learned feature representations. Most feature engineering practices in modern CV literature involve automatic learning of feature extractors via deep learning. We describe this family of techniques in the next section.

*Representation Learning via Meta-Learning and Self-Supervised Learning*

While carefully and manually crafted feature descriptors and feature extractors can work well in some domains, it is often nontrivial to develop such representations for social human behavior. Representation learning, or the automated learning of useful feature representations, has become a feasible approach with the recent democratization of deep neural network training capabilities and widespread availability of well-equipped graphics processing units (GPUs).

Traditional transfer learning approaches require extensive labeled training data to learn the source representation. In some cases, pretrained models with relevant features are unavailable, and there are not enough labeled images in the target domain – even with data augmentation – to warrant the use of a CNN. Human-engineered feature extraction methods may be unavailable for the domain of interest. Most types of human social behavior fall under this category, as there are few public datasets with *labeled* social behaviors. Self-supervised learning was designed for these situations. Self-supervised learning consists of training a classifier in a task that can be learned without labels. In natural language processing, common self-supervised training tasks include (1) masked language modeling, where certain words in a text corpus are randomly removed and the classifier must predict the missing words, and (2) next sentence prediction, where the goal is to predict whether, given two sentences, the second sentences immediately follows the first sentence. The Bidirectional Encoder Representations of Transformers (BERT) model was popularly pretrained using these two methods [50], and pretrained BERT models and newer variants are able to outperform other language tasks such as question answering, named entity recognition, and sequence classification, among many others. Approaches in CV include rotating an image and predicting the rotation angle [75, 222], splitting an image into smaller segments and solving a jigsaw puzzle from the shuffled segments [146], predicting the relative spatial location of two segments of the same image [52], and using triplet loss between an image, a manipulated version of the image using one or more data augmentation strategies, and a separate source image [54]. While the performance of different semi-supervised training tasks will vary across target domains, the idea is that the source tasks learn fundamental properties of the images which can transfer to a variety of prediction tasks.

Finding good source datasets and corresponding self-supervised approaches for social human behavior is a difficult and relatively unexplored area of research. Self-supervised methods for motion and pose related features have been proposed. 3D motion tracking from a monocular camera was improved by self-supervising on high-performing 2D CV tasks [192]. In particular, self-supervision on 2D keypoint detections learned features useful for 3D body joint predictions, self-supervision on 2D optical flow vectors learned features useful for 3D mesh vertex displacement between frames, and self-supervision on 2D figure-ground segmentation learned features useful for 3D mesh prediction. Other approaches leverage the output of well-established ML methods. For example, 3D pose can be estimated by combining 2D pose estimates from different camera angles using epipolar geometry and then training a 3D pose estimator using the inferred 3D pose information [104]. The majority of self-supervised methods for human behavior are focused on activity recognition based on movement and positional features.

Representation learning for social behavior is ill-studied, especially for CV. Transfer learning over CNNs trained on ImageNet features has been applied to predicting depression and aggression related to dementia [215]. ImageNet features are most useful for object recognition rather than complex affect prediction. Some datasets, such as popular affective computing datasets like the Extended Cohn-Kanade dataset [127] and the Facial Expression Recogition 2013 dataset [76], can more effectively transfer into other affective tasks [141]. Self-supervised learning approaches may often focus on the face, which provides much detail about social intention. For example, self-supervised learning of facial dynamics in videos has been shown to learn baseline model weights useful for personality prediction [178].

Self-supervised learning techniques which are specialized towards more complex nonverbal social cues are virtually nonexistent to date. ML engineers must employ the techniques described here to learn what is hopefully a useful representation for transfer learning. Like in other domains, however, the most effective techniques will require creativity in designing the source task.

Meta-learning has emerged an area of study for pretraining models so that they can quickly learn in new domains. We describe meta-learning efforts on CV for emotion recognition, as human affect is a prototypical nonverbal communicative behavior which the ML research community has exerted much effort towards. Meta-learning has been applied to personalized facial action unit detection for emotion recognition [111]. Action units are a popular feature representation for emotion and cognition recognition, but action unit detection notoriously struggles to adapt to unfamiliar settings such as new human subjects. Model-agnostic meta-learning was applied to maximize model parameters to efficiently learn across a variety of action units and subjects. Prototypical learning, where the "prototypical" representation of each class, has been explored on popular emotion recognition datasets [109].

Meta-learning methods rely on a "support set" consisting of several representative datasets which are used to learn initial model parameters prior to the "query set", which is the target dataset. At the time of writing, however, there are few representative datasets for most nonverbal social behaviors, thereby limiting the potential for widespread research into meta-learning in these domains. We call for researchers and practitioners to release public datasets of social human behavior, many of which are certainly curated and labeled by individual institutions but not shared with the broader public due to competition.

*Active Learning*

Active learning, or the use of metrics to estimate which unlabeled data points would provide the greatest performance boost to a classifier if those points were to be labeled, is an active area of ML research. There are two major paradigms in active learning literature: uncertainty sampling and diversity sampling. In uncertainty sampling, points are labeled based on the classifier's output probability confidence about unlabeled data points. In diversity sampling, points are selected based on some measure of similarity between the points. Another method is query by committee, where multiple classifiers vote on the expected output of an unlabeled data point, and points with maximal disagreement are labeled. More complex methods have been proposed to attempt to balance the exploration-exploitation tradeoff [23], such as Active Thomas Sampling [26]. Other methods leverage the inner workings of the model in question, such as the minimum marginal hyperplane technique used with support vector machines, where data points with the minimum difference from the classification separation boundary are prioritized.

A major challenge in uncertainty sampling for CV deep learning models is that traditional CNNs do not contain properties amenable to active learning. Unlike simple models such as logistic regression, CNNs are uncalibrated, meaning that their output probability distribution in the final softmax or sigmoid network layer does not represent confidence well, minimizing the impact of uncertainty sampling methods. Traditional CNNs happen to be overconfident. Bayesian neural networks have been proposed as a solution to the calibration issue by sampling each network parameter from a distribution [69]. During network inference, all possible parameters are integrated over, simulating an infinite ensemble of neural networks, thus providing a better uncertainty measure. A challenge is that Bayesian CNNs tend to underperform compared to traditional CNNs, thus requiring training of two separate model architectures, one for active learning and one for inference. Other approaches to uncertainty sampling for deep neural networks include learning the loss function directly [228].

Diversity sampling methods as well as methods which combine uncertainty sampling and diversity sampling have also been proposed. A diversity sampling approach used in deep neural networks for CV is to find the "core set" of images which maximize the variety of image features in the dataset and are thus representative of the entire dataset [172]. The primary challenge for this approach is finding an image representation which captures the features of importance to social behavior without capturing the irrelevant features. Methods which combine diversity and uncertainty sampling include active learning acquisition functions which consider both the mutual information between model output and parameters as well as similarity between images [101], using weighted k-means clustering based on intermediate network embedding and weighted by uncertainty metrics [239], and computing a gradient vector of the model output with respect to model weights which can be clustered to find similar points [11].

The overarching challenge in active learning research is that no single metric-based strategy, whether based on uncertainty or diversity, will work for all types of data. For example, datasets may consist of several repetitive data points, resulting in redundancy when using a static metric such as maximum entropy. Adaptive strategies modeled after reinforcement learning can learn a policy for selecting salient data points, thereby "learning how to learn" [64]. In some cases tailored to CV for human behavior, the active learning system relies on a policy function which is personalized for each human subject [165]. Reinforcement learning for active learning is a relatively understudied area, but initial approaches rely on the formulation of active learning as a Markov Decision Process (MDP). Some approaches model active learning as a streaming process, where individual frames are sent to the system, which then decides whether to label the data point [64, 165]. Other formulations select an individual point from a large pool [120, 200]. The reward of the MDP is usually a function of the increase in classifier performance according to one or more metrics when new data are added [64, 120, 152, 165]. We note that this reward measurement in each state of the MDP is nontrivial, as it requires heavy computation per iteration of active learning to retrain the classifier and measure its performance. The MDP state sometimes does not account for the discrete time inherent of a MDP and is modeled as the point or points being considered or some function (feature representation) of the point [165, 217]. Because uncertainty-based active learning requires a metric for classifier uncertainty, the state can also include the classifier's parameters or even the prediction itself [152], which is a function of the data point. The MDP action is usually the choice of one or more unlabeled points to label [64, 165].

While reinforcement learning and other adaptive active learning strategies have been validated from a research context, training a policy function requires vast amounts of (state, action, reward) tuples and extensive compute power to train, and the acquisition of the reward is in itself expensive since it requires retraining a classifier. Given this resource constraint and the significant thought and engineering overhead, adaptive active learning may not be practically useful given current practices. However, if more efficient MDP formulations and representations of state, action, and reward are developed, then efficient labeling of the massive, high dimensional, complex, and repetitive nature of the datasets may become widespread.

In lieu of or in addition to active learning, automatic labeling of data points via semi-supervised learning can considerably reduce the required number of human labels, making sufficient labeled coverage of the dataset tractable. Semi-supervised learning involves pseudo-labels which are generated by the classifier for unlabeled data points. Semi-supervised learning works best when the distributions of properties of the unlabeled points are similar to those of the labeled points used to train the classifier. Unfortunately, because social behavior classifiers are often trained on data from naturalistic settings, a pretrained model's predictions on a new image from a new individual in a new setting is particularly prone to wrong predictions due to overfitting.

*Generative Data Augmentation*

Data augmentation is an approach for handling noisy data streams. Data augmentation consists of generating modifications to the input dataset which maintain the original categorization of the data but provide the classifier examples of the data point in different contexts. Examples of common image-based data augmentations include adjusting brightness, zooming, flipping, rotating, injecting noise, and cropping. Recently, generative adversarial networks (GANs) have been used to generate completely synthetic images, including for social behavior tasks such as emotion recognition [227]. Interpolation on the latent space learned by variational autoencoders has also enabled non-trivial data augmentations [119].

Generative models, through conditioning, are able to perform "domain transfer", a form of image manipulation where certain features of the image are preserved while others are converted into the desired domain. This approach is often used in the machine learning literature to generate synthetic samples to render model adaptation possible with fewer samples. Wang et al. demonstrate the use of synthetic images in image classification settings [202]. Niinuma et al. [144] and Zhu et al. [241] show the possible performance boosts that can be gained by generating synthetic facial expressions with a GAN. Similar uses of generative models can be applied to different behavioral phenotyping paradigms such as headbanging detection and voice related classifications.

Data augmentation is also often used as a solution for biased datasets. Sometimes, differences in number, quality, and diversity of data samples across demographic groups are inevitable. A potential mitigator is to turn to data augmentation to equalize noise across classes. Data augmentation has been shown to reduce dataset bias [188] when applying background substitutions, geometric transformations such as translation, mirroring, and rotation, and color transformations such as brightness modification and color channel intensity modifications [132]. GANs have been used to perform targeted data augmentation by transferring sensitive properties of data points between groups. For example, racial characteristics have been transferred across images [231] using CycleGANs, which enable image-to-image translation when paired examples are not present by using a combination of adversarial loss to ensure generated images in the target domain are consistent with real images in the target domain along with cycle consistency loss to maximize the reconstruction of an image when domain transfers between both domains have been consecutively applied [240]. GANs and variational autoencoders (VAEs) have been combined in a VAE-GAN architecture to reduce dataset bias by enforcing a shared latent space which is used both for image reconstruction and domain transfer. The end of the VAE also serves as the generator network in a GAN. The discriminator evaluates whether the domain transferred images are realistic [94, 117]. Adversarial learning can also minimize bias by using a discriminator to predict a sensitive latent variable known in advance, such as a demographic group [234].

*Label Representations*

Datasets often contain subjective data labels. One solution to handle this in ML is to explicitly encode the ambiguity in the labels. Soft-target labeling is the use of a probabilistic label vector in lieu of traditional one-hot encoded vectors. Using a traditional loss function such as cross-entropy loss with soft-target labels will promote learning of an output vector which mimics the distribution used to train the classifier. Soft-target labels have been successfully deployed in emotion recognition models [8, 83, 207, 246], pedestrian pose orientation [85], and mispronunciation detection in audio-based tone models [115].

The challenge with training using soft-target labels arises when the probability distribution is unknown in advance. Crowdsourcing has been used as a possible solution, where several annotations are acquired per data point [125, 246]. Active learning with soft target labels has been proposed and prototyped for soft target labels by the authors of this review (this work is currently under preparation). In this scenario, each image receives additional labels until the entropy of the crowd label distribution is sufficiently low. Active learning for crowdsourced soft-target labels is an understudied area of ML research open to much innovation.

Creating appropriate label representations is also important when handling classification of multiple behaviors within a single image. Multi-label classification is a well-studied area of ML research where the goal is to predict one or more class memberships for each data point. One common approach is modeling the multi-class problem as several independent classifiers or as a chain of connected classifiers where the output of a classifier in a sequence of binary classifiers is provided as input to the next classifier [160]. The difficulty with this approach is that CNNs are expensive to train, and training several binary classifiers is prohibitively expensive. This idea has been extended to recurrent neural networks (RNNs) for classification of multiple objects in images, where only a single CNN is used to extract features of the image in conjunction with a label embedding to predict subsequent labels [199]. Another approach is to label each combination of classes as a single class while otherwise modeling the classification task as a traditional multi-class problem. This approach can work well with a small number of potential labels, but quickly becomes intractable with more than a few possible labels. Ensemble methods have also been used, but a limiting challenge for images becomes the infeasibility of training multiple CNNs and the lack of sufficiently unique feature representations when using lightweight models.

*Crowd Worker Selection*

The complexity of behavioral imaging data usually requires high quality annotators. On traditional crowdsourced data labeling tasks, the data points to label are instantly identifiable by nonexperts, and the correct label is not subjective or ambiguous. For example, ImageNet [48], the most popular CV dataset for object recognition, consists of *nouns* such as "car mirror", "car wheel", "hourglass", and "printer". By contrast, classification of human behavior requires understanding of *verbs*. Activity recognition, a popular subfield of CV, includes datasets such as the Kinetics 400 Dataset [100] with classes such as "kissing", "hugging", "ice skating", and "playing guitar" - most or all of which are unambiguous. Labeling of clinically useful behaviors for behavioral and psychiatric diagnoses, however, is much more complex. The labels can be subjective, with multiple correct labels or even multiple labels at once comprising *the* correct label. Oftentimes, domain experts are required to provide reliable labels, but this is not scalable in practice. One solution is to train a crowd workforce to distinguish behaviors of interest. The limiting issue of this approach is that social behavior is particularly complex and nuanced and is therefore challenging to reliably train for. An alternative approach is to curate a skilled workforce of "super recognizers" who consistently annotate complex social behaviors correctly in a series of test tasks [204-209].

When crowdsourcing, it is critical to use a platform which provides access to an unrestricted global worker pool, such as Microworkers [86], which allows task creators to specify which country or countries the workers must be based in to complete the work. Notably, MTurk does not provide such functionality at the time of this writing. There is room for more sophisticated crowdsourcing platforms which provide granular demographic and performance information about each crowdworker anonymously, thereby providing ML engineers with increased confidence about the quality and lack of biases of their datasets' labels.

Industries centered around data labeling for AI have risen as a natural byproduct of the ease of ML model development and training. Crowdsourcing services such as Amazon Mechanical Turk (MTurk) [153] and Microworkers [86] provide organizations with a large pool of online gig workers who can provide data labels as a service. Other enterprises, such as Hive, CloudFactory, and Humans in the Loop, provide management of large data labeling workforces who can be trained according to a client's requirements. Further explorations into the methods described in this section could enable commercial data labeling services to expand to more complex, high dimensional, and noisy datasets. Sophisticated data labeling is a nascent subfield of data science, and it is likely that advancements in efficient labeling assignment algorithms will propel model development efforts.

When crowdsourcing labels for images depicting human behaviors, privacy is a core concern. A recently adopted approach to privacy in the labeling process which does not require obfuscation of the data is the adoption of a private crowd of annotators who are vetted and trusted with seeing PHI data. While crowd workers on platforms like MTurk and Microworkers cannot be blindly trusted, we have in previous work demonstrated that diagnostic tasks for behavioral conditions like pediatric autism can be tagged by privacy crowd workers who have engendered trust by researchers through repeatedly high performance on a subset of representative public datasets. These workers completed HIPAA training, CITI training, and encrypted their laptop with university encryption standards to become official members of the research team [209]. Several strategies have been tested for the selection of trustworthy crowd workers, including behavioral metrics of the crowd worker such as the consistency of workers [205] and estimations of the correct class by a pretrained model [206]. Crowdsourcing metrics for analyzing worker trustworthiness have been largely explored in human-computer interaction literature, including peer review by other crowdworkers, self assessment, comparison against gold standards, and allowing workers to view other workers' submissions to modify their answers (known as "recombination") [56]. Further checks on workers, such as background checks and more extensive training, may be necessary for especially sensitive images.

*Federated Learning*

Another general-purpose privacy preserving paradigm is federated learning. In federated learning, a model is locally trained on a device or system which stores the training data for a single user or family. To train a general-purpose model, only model coefficients are uploaded to a central server, which aggregates each individual's learned set of parameters to create a global model. Because federated learning often consists of smaller training sets at each local node, strong feature representations and pretraining must be considered for complex image data. A major challenge in federated learning arises from data distributions which are not independent and identically distributed (IID). Possible causes of non-IID data with respect to human behavior include natural differences in feature distributions across individuals (e.g., different baseline facial expression tendencies), difference label frequencies (e.g., different tendencies towards making certain facial expressions), and different amounts of data across users. Approaches to account for non-IID data include data augmentation across clients, creating custom loss functions to handle the unequal distributions, and personalization of models for each user, an approach which reframes the non-IID nature of the data as a "feature not a bug" [95]. We discuss personalization in more detail in the next section.

**Discussion and Conclusion**

ML solutions which can precisely and reliably measure human behavior from rich multimedia data streams are within reach. However, additional creativity in both data processing and CV methods are required before high-performing classifiers which can be used in commercial and clinical contexts become feasible. Deep neural networks contain the capacity to approximate any function, including the blatantly nonlinear and complex patterns required to discern complex context-dependent expressions and actions. Modern GPUs have the capacity to train these networks in a tractable amount of time. Some of the largest challenges and opportunities for digital phenotyping with CV fall earlier in the model development pipeline: data acquisition, cleaning, and labeling. Algorithmic innovations in these areas, which are often overlooked by ML researchers, in conjunction with additional creativity when applying state-of-the-art CV feature learning, data augmentation, and supervised approaches, can sufficiently push the needle forward enough to enable deployable CV models of social behavior.

The outlook for CV-based digital phenotyping is promising. However, current efforts are mostly focused on the second half of the pipeline, starting at the feature extraction step. Because ML researchers often consider innovations in model development to be the "sexiest", the necessary sophistication of data curation, labeling, and quality control algorithms is lacking, therefore limiting the full potential of advanced ML techniques and preventing advancements in downstream model development efforts which would naturally arise from having higher quality and more complex datasets to work with. We conclude with a call-to-action for algorithmic research into novel data acquisition and management pipelines which address the challenges and opportunities discussed here. Such innovations will enable novel practical applications to ML as a whole, including complex CV tasks such as quantifying human behavior.

Unlike other domains where data are abundant and easy for an untrained individual to label, computational psychiatry requires the handling of especially noisy and heterogeneous data streams, working with ambiguous and compound class memberships, addressing of user privacy concerns, producing creative representation learning approaches, and development of personalized models. All these issues require additional innovations before CV for social human behavior, with its myriad potential applications in commercial and clinical settings, can be realized.

**Acknowledgements**

This work was supported in part by funds to DPW from the National Institutes of Health (1R01EB025025-01, 1R21HD091500- 01, 1R01LM013083), the National Science Foundation (Award 2014232), The Hartwell Foundation, Bill and Melinda Gates Foundation, Coulter Foundation, Lucile Packard Foundation, the Weston Havens Foundation, and program grants from Stanford's Human Centered Artificial Intelligence Program, Stanford's Precision Health and Integrated Diagnostics Center (PHIND), Stanford's Beckman Center, Stanford's Bio-X Center, Predictives and Diagnostics Accelerator (SPADA) Spectrum, Stanford's Spark Program in Translational Research, and from Stanford's Wu Tsai Neurosciences Institute's Neuroscience: Translate Program. We also acknowledge generous support from David Orr, Imma Calvo, Bobby Dekesyer and Peter Sullivan. P.W. would like to acknowledge support from Mr. Schroeder and the Stanford Interdisciplinary Graduate Fellowship (SIGF) as the Schroeder Family Goldman Sachs Graduate Fellow.

163. Ekman, Paul, and Erika L. Rosenberg, eds. *What the face reveals: Basic and applied studies of spontaneous expression using the Facial Action Coding System (FACS)*. Oxford University Press, USA, 1997.
164. Rouast, Philipp Vincent, Marc Thomas Philipp Adam, Tracy Burrows, Raymond Chiong, and M. E. Rollo. "Using Deep Learning and 360 Video to Detect Eating Behavior for User Assistance Systems." In *ECIS*, p. 101. 2018.
165. Rudovic, Ognjen, Hae Won Park, John Busche, Björn Schuller, Cynthia Breazeal, and Rosalind W. Picard. "Personalized estimation of engagement from videos using active learning with deep reinforcement learning." In *2019 IEEE/CVF Conference on CV and Pattern Recognition Workshops (CVPRW)*, pp. 217-226. IEEE, 2019.
166. Safdar, Saba, Wolfgang Friedlmeier, David Matsumoto, Seung Hee Yoo, Catherine T. Kwantes, Hisako Kakai, and Eri Shigemasu. "Variations of emotional display rules within and across cultures: A comparison between Canada, USA, and Japan." *Canadian Journal of Behavioural Science/Revue canadienne des sciences du comportement* 41, no. 1 (2009): 1.
167. Sage, Kingsley, and Stewart Young. "Security applications of CV." *IEEE aerospace and electronic systems magazine* 14, no. 4 (1999): 19-29.
168. Sánchez-Lozano, Enrique, Georgios Tzimiropoulos, Brais Martinez, Fernando De la Torre, and Michel Valstar. "A functional regression approach to facial landmark tracking." *IEEE transactions on pattern analysis and machine intelligence* 40, no. 9 (2017): 2037-2050.
169. Sapiro, Guillermo, Jordan Hashemi, and Geraldine Dawson. "CV and behavioral phenotyping: an autism case study." *Current Opinion in Biomedical Engineering* 9 (2019): 14-20.
170. Saponaro, Philip, Haoran Wei, Gregory Dominick, and Chandra Kambhamettu. "Estimating Physical Activity Intensity And Energy Expenditure Using CV On Videos." In *2019 IEEE International Conference on Image Processing (ICIP)*, pp. 3631-3635. IEEE, 2019.
171. Schuch, Kelsey N., Lakshmi Narasimhan Govindarajan, Yuliang Guo, Saba N. Baskoylu, Sarah Kim, Benjamin Kimia, Thomas Serre, and Anne C. Hart. "Discriminating between sleep and exercise-induced fatigue using CV and behavioral genetics." *Journal of Neurogenetics* 34, no. 3-4 (2020): 453-465.
172. Sener, Ozan, and Silvio Savarese. "Active learning for convolutional neural networks: A core-set approach." *arXiv preprint arXiv:1708.00489* (2017).
173. Shen, Jie, Stefanos Zafeiriou, Grigoris G. Chrysos, Jean Kossaifi, Georgios Tzimiropoulos, and Maja Pantic. "The first facial landmark tracking in-the-wild challenge: Benchmark and results." In *Proceedings of the IEEE international conference on CV workshops*, pp. 50-58. 2015.
174. Shi, Yong, Zhao Lv, Ning Bi, and Chao Zhang. "An improved SIFT algorithm for robust emotion recognition under various face poses and illuminations." *Neural Computing and Applications* 32, no. 13 (2020): 9267-9281.
175. Simon, Tomas, Hanbyul Joo, Iain Matthews, and Yaser Sheikh. "Hand keypoint detection in single images using multiview bootstrapping." In *Proceedings of the IEEE conference on CV and Pattern Recognition*, pp. 1145-1153. 2017.
176. Singletary, Bradley A., and Thad E. Starner. "Learning visual models of social engagement." In *Proceedings IEEE ICCV Workshop on Recognition, Analysis, and Tracking of Faces and Gestures in Real-Time Systems*, pp. 141-148. IEEE, 2001.
177. Sirisuriya, De S. "A comparative study on web scraping." (2015).
178. Song, Siyang, Shashank Jaiswal, Enrique Sanchez, Georgios Tzimiropoulos, Linlin Shen, and Michel Valstar. "Self-supervised Learning of Person-specific Facial Dynamics for Automatic Personality Recognition." *IEEE Transactions on Affective Computing* (2021).
179. Soomro, Khurram, Amir Roshan Zamir, and Mubarak Shah. "UCF101: A dataset of 101 human actions classes from videos in the wild." *arXiv preprint arXiv:1212.0402* (2012).
180. Stark, David E., Rajiv B. Kumar, Christopher A. Longhurst, and Dennis P. Wall. "The quantified brain: a framework for mobile device-based assessment of behavior and neurological function." *Applied clinical informatics* 7, no. 02 (2016): 290-298.